\documentclass{article}

\usepackage[final]{neurips_2024}

\usepackage{graphicx}
\usepackage{epstopdf} 
\usepackage{subcaption}

\usepackage[utf8]{inputenc} 
\usepackage[T1]{fontenc}    
\usepackage{hyperref}       
\usepackage{url}            
\usepackage{booktabs}       
\usepackage{amsfonts}       
\usepackage{nicefrac}       
\usepackage{microtype}      
\usepackage{xcolor}         
\usepackage{subcaption} 

\title{Agentic-HLS: An agentic reasoning based high-level synthesis system using large language models \newline (AI for EDA workshop 2024)}

\author{%
Ali Emre Oztas \quad Mahdi Jelodari\thanks{visiting scholar during the course of this research and contest.} \\
Department of Computer Science,\\
The University of Manchester \\
Manchester, UK\\
\texttt{\{aliemre.oztas,mahdi.jelodari\}@manchester.ac.uk}
}

\begin{document}

\maketitle

\begin{abstract}

Our aim for the ML Contest for Chip Design with HLS 2024 was to predict the validity, running latency in the form of cycle counts, utilization rate of BRAM (util-BRAM), utilization rate of lookup tables (uti-LUT), utilization rate of flip flops (util-FF), and the utilization rate of digital signal processors (util-DSP). We used Chain-of-thought techniques with large language models to perform classification and regression tasks. Our prediction is that with larger models reasoning was much improved. We release our prompts and propose a HLS benchmarking task for LLMs.

\end{abstract}

\section{Introduction}

High-level synthesis (HLS) has revolutionised the landscape of hardware design by raising the level of abstraction, making it feasible to develop domain-specific accelerators (DSAs), such as field-programmable gate arrays (FPGAs), using high-level programming languages like C/C++ instead of traditional hardware description languages (HDLs). The figure to the right illustrates a typical HLS design, where the functionality is described using C++ code, and performance-related compiler directives, known as pragmas, dictate the microarchitectural transformations within the HLS framework. These pragmas influence key performance metrics, including latency and resource utilization rates.

Despite these advancements, the optimisation space for microarchitectures expands exponentially with the inclusion of pragmas, creating an overwhelming design space. Evaluating each candidate architecture through HLS tools requires substantial time—ranging from several minutes to hours—making the optimization process both labor-intensive and time-consuming. Machine learning models have been leveraged to expedite this process, allowing for design quality predictions in milliseconds [2,3].

In the ML Contest for Chip Design with HLS [8], we used a novel approach that introduces agentic high-level synthesis, leveraging large language models (LLMs) equipped with specialised tools for reasoning and optimisation. By imbuing these models, specifically over 100 billion parameter models, with agent-like capabilities, they can autonomously explore, assess, and optimise HLS designs, enhancing the automation and intelligence embedded in the design flow. Our method bridges machine learning and hardware design, empowering LLMs to understand and interact with the design space to produce superior configurations more efficiently.

\section{Methodology}

\begin{figure}
  \centering
  \label{system-figure}  
  \includegraphics[totalheight=8cm]{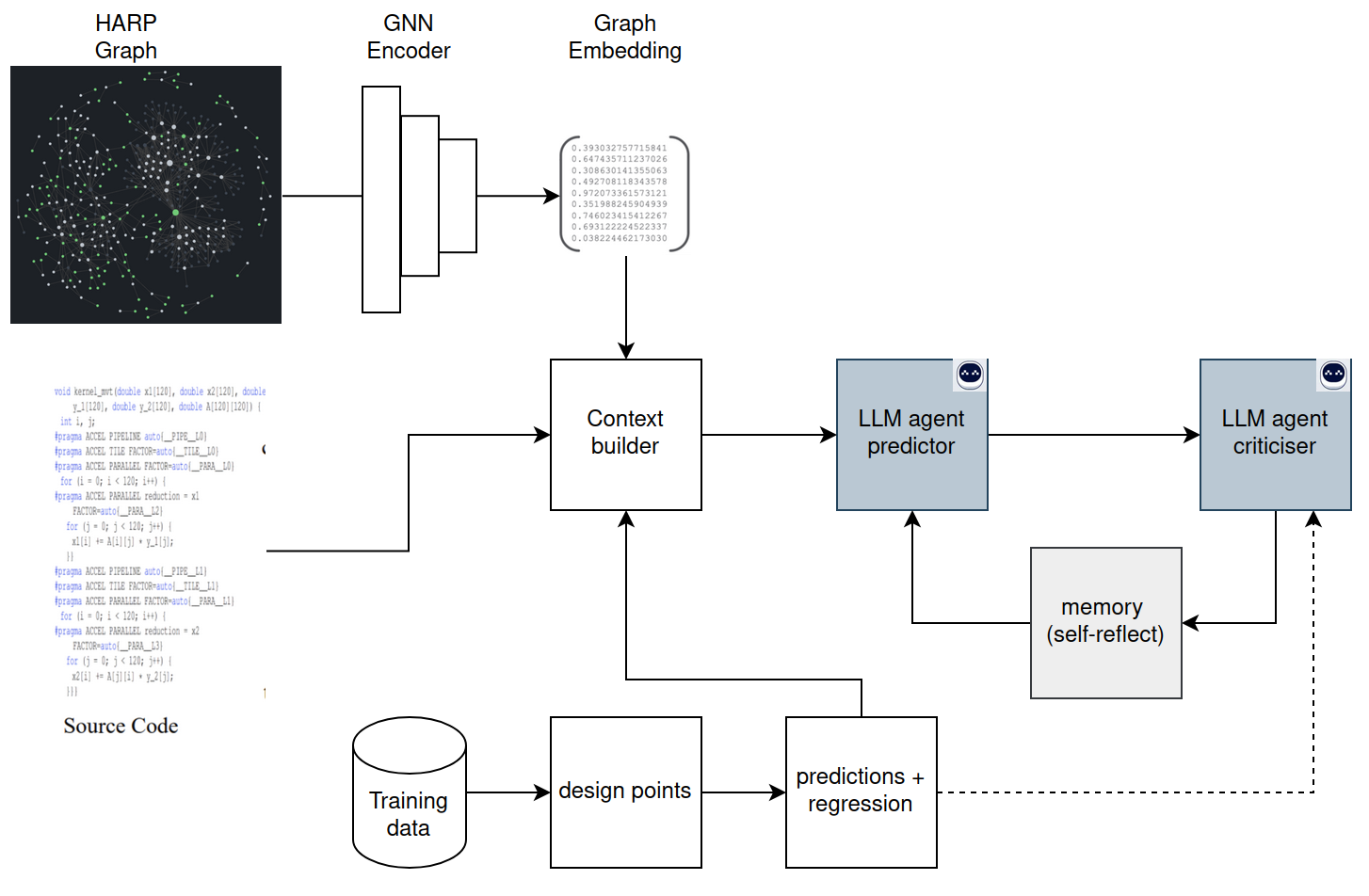}
  \caption{Agentic-HLS system architecture, a LLM+GNN system that uses agentic reasoning to iteratively review its output and refine its predictions.}
  \label{fig-system}
\end{figure}

Our proposed approach for the contest relies on an extended informative representation of an input design for high-quality performance prediction shown in Figure~\ref{fig-system}. Our approach consists of several steps: (a) HARP fine-tuning, (b) source code sequence analysis, (c) reasoning steps based on the training data points, and (d) an agentic predictor/criticiser to evaluate the results.

\subsection{Fine-tuning HARP}
HARP [3] is a GNN-based methodology that enhances FPGA HLS optimization by using a hierarchical graph representation to model program structure and pragma transformations separately, allowing for improved accuracy and adaptability. Through its modular design, HARP efficiently captures the dynamic impact of design choices, enabling faster and more accurate design space exploration. We utilized the HARP encoder to create comprehensive graph embeddings that encapsulate critical design features. The graph embeddings provide a representation from which reasoning about the critical path and latency measurements can be derived. The structure of these embeddings facilitates an in-depth analysis that highlights potential performance bottlenecks.

\subsection{Source Code Analysis}
\begin{figure}
  \centering
  \label{fig:source_analysis2}
  \includegraphics[totalheight=6cm]{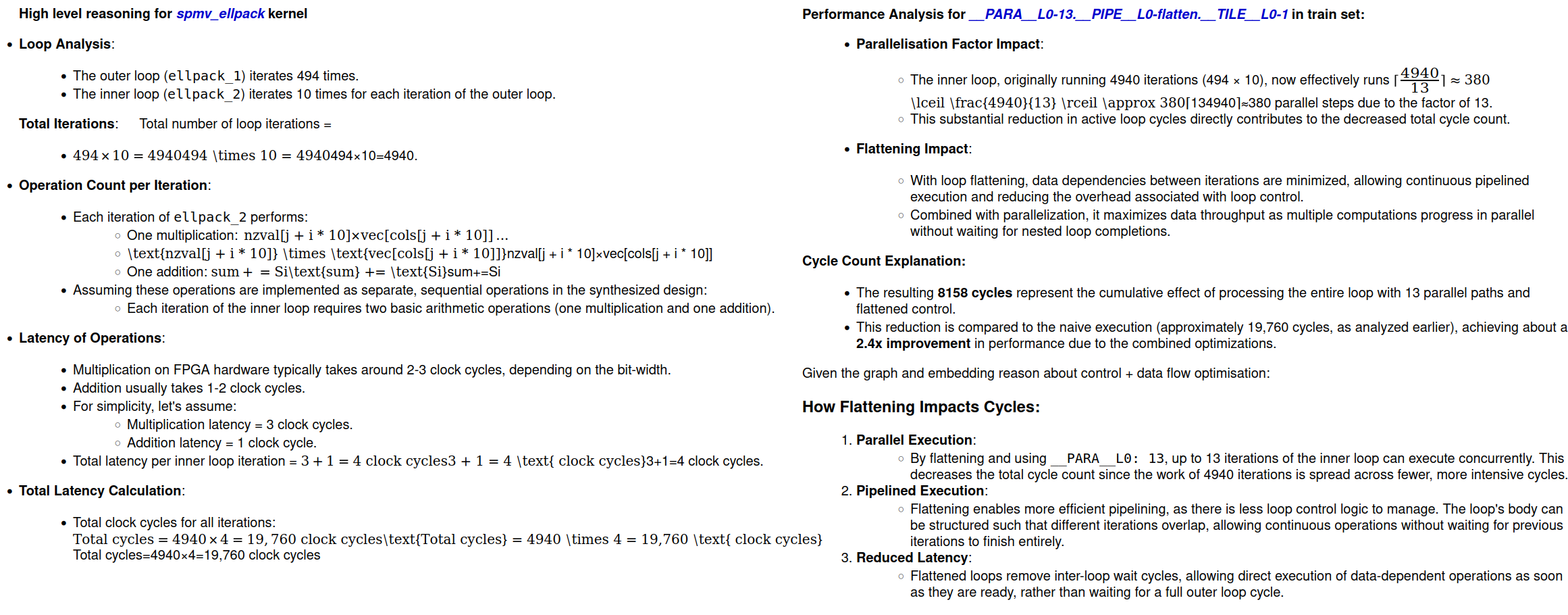}
  \caption{High level reasoning of the target kernel (left) and pragma impact on the control and data flow of the kernel generated by GPT4o (right).}
  \label{fig}
\end{figure}

The source code is analyzed using a large language model (LLM) to identify key program structures such as blocks and loops, aiding in the construction of data and control flow graphs. As illustrated in Figure ~\ref{fig}, the LLM-generated analysis serves as an initial step prior to further optimizations. 

\subsubsection{Reasoning}
\begin{figure}[htbp]
  \centering
  \begin{subfigure}{0.32\textwidth} 
    \centering
    \includegraphics[width=1.1\linewidth]{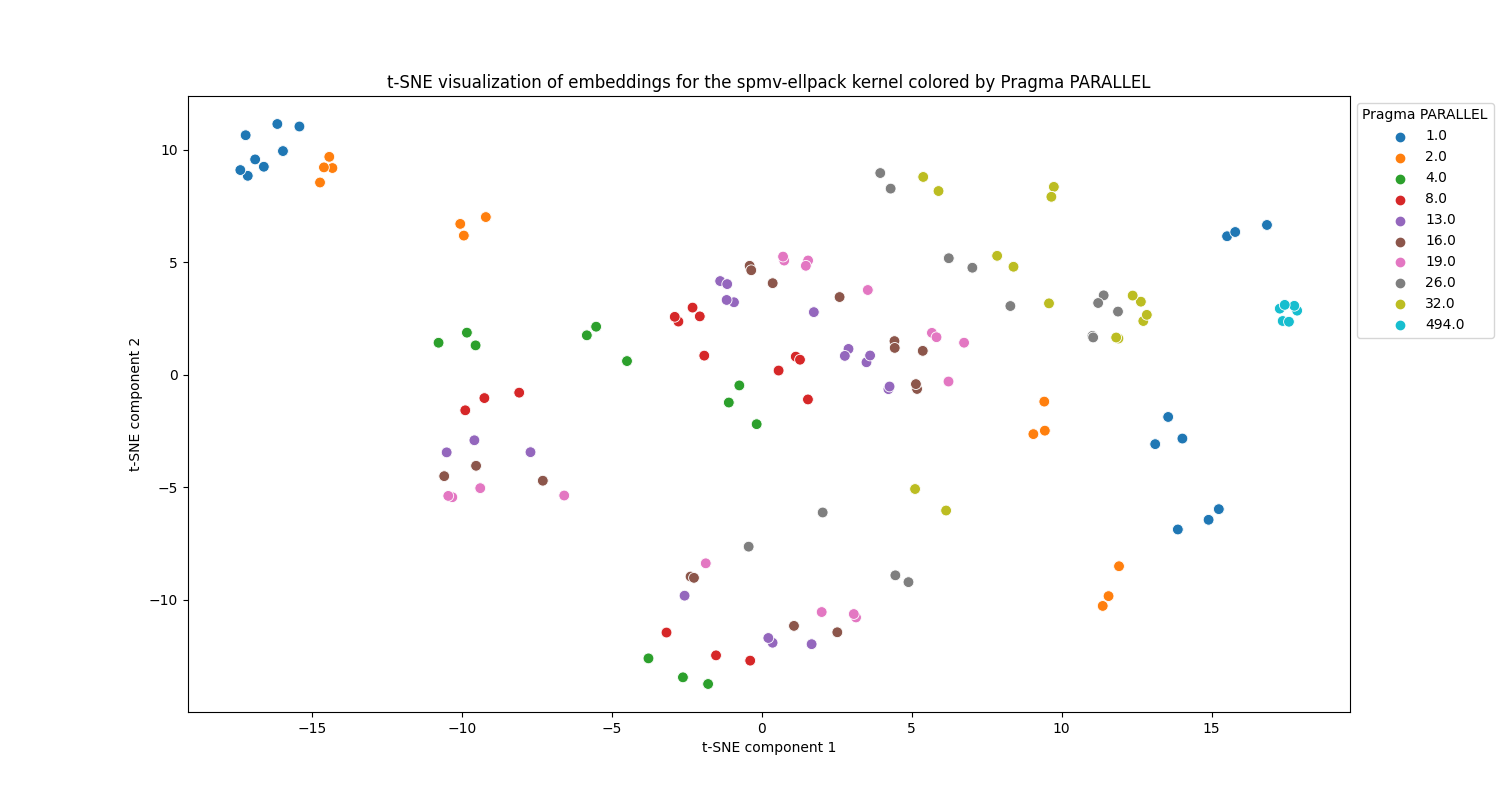} 
    \subcaption{}
  \end{subfigure}
  \hfill 
  \begin{subfigure}{0.32\textwidth}
    \centering
    \includegraphics[width=1.1\linewidth]{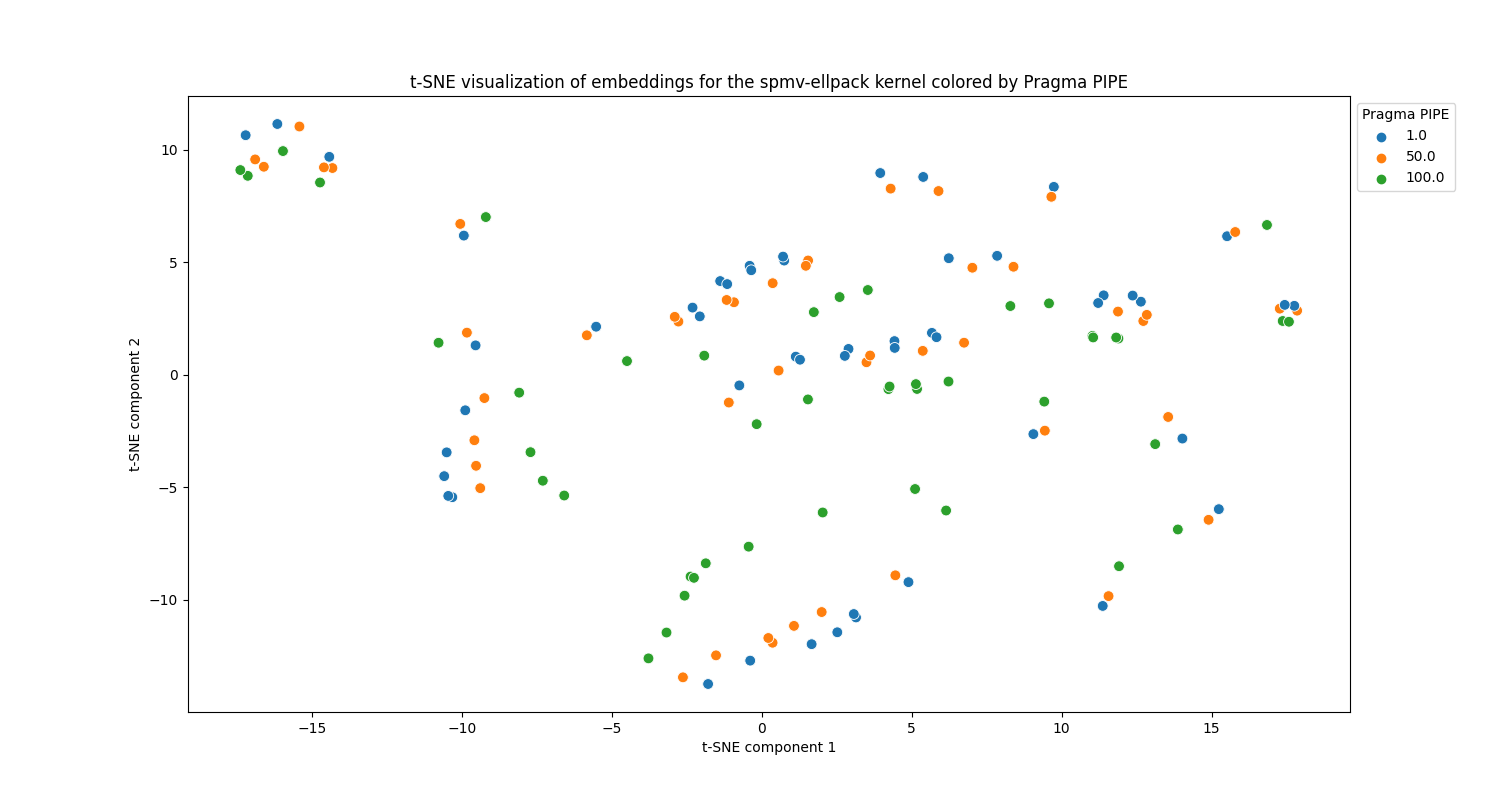}
    \subcaption{}
  \end{subfigure}
  \hfill 
  \begin{subfigure}{0.32\textwidth}
    \centering
    \includegraphics[width=1.1\linewidth]{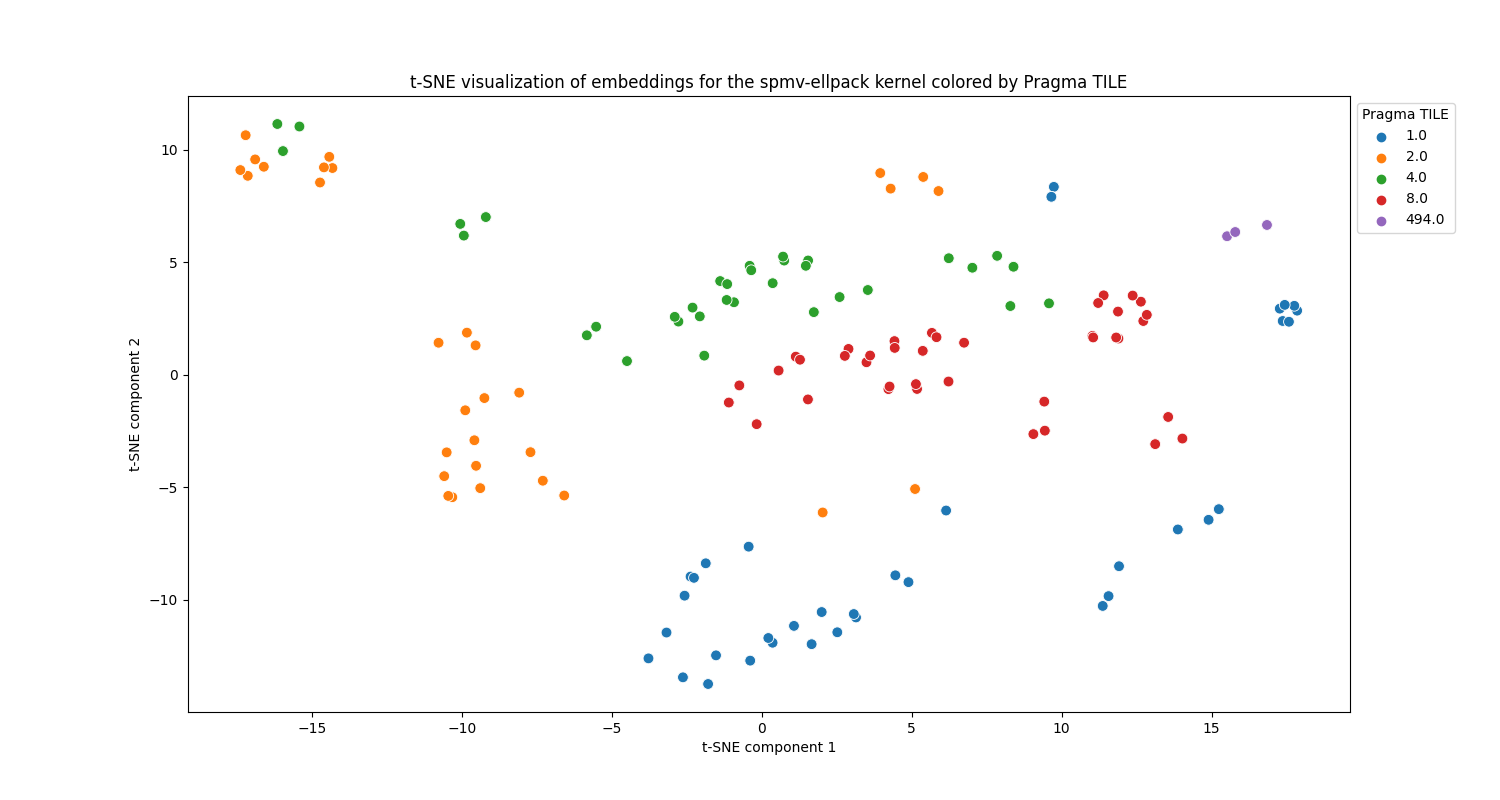}
    \subcaption{}
  \end{subfigure}
  
  \caption{t-SNE representations of graph embeddings for the spmv-ellpack kernel with different Pragma values: (a) PARALLEL, (b) PIPE, (c) TILE. }
  \label{fig:tsne_embeddings}
\end{figure}

We provide a high-level explanation of the impact that the target pragmas may have on the control/data flow graphs. The model incorporates graph-level reasoning to assess the influence of each pragma and its effect on the design's critical path and validity. This capability enables the prediction model to evaluate potential performance variations induced by different pragmas. Figure~\ref{fig:tsne_embeddings} shows impact of various optimisations on the spmv-ellpack kernel. Outliers were mostly identified as invalid designs. Graph embeddings carry specific control/data flow information that enables the LLM to attend to node-edge relationships and enhances their overall reasoning.

\subsubsection{Agentic Evaluation}
An iterative evaluation process is employed, wherein the predictor agent performs self-reflection based on available training data batches. This process runs for three cycles due to its computational expense but is crucial for enabling the predictor to rationalize its performance predictions. The predictor-agent iteratively refines its analysis by providing justifications for specific optimizations and then reflecting on feedback from the criticiser agent, enhancing overall prediction reliability.

\section{Results}

We initially began with a solution entirely based on large language models (LLMs), employing GPT-4o solely for generating predictions. Preliminary evaluations indicated that the classification task (i.e., distinguishing valid from invalid designs) significantly affected the root mean square error (RMSE) score. Consequently, our primary focus shifted towards developing a highly accurate classifier. Concurrently, we explored the design space exploration (GNN-DSE) method and deployed a local LLaMA2 7B model for classification. However, this model lacked sufficient depth for complex reasoning tasks.

Subsequently, we concentrated on running HARP locally and contributed a pull request (PR) to resolve several bugs, enabling us to fine-tune HARP on the training dataset. We leveraged the reasoning capabilities provided by HARP graph embeddings and combined them with source code sequencing, which yielded a notable improvement over LLaMA2 (27\%). Our final solution integrated GPT-4o and included several reasoning components along with an evaluator agent to iteratively (3 iterations) assess the predicted outcomes using the training samples available to the model. The results are detailed in Table \ref{table-1}. For the final submission, we selected the best-performing models, including the fine-tuned HARP.

The code repository will be updated: https://github.com/Zheyu-Rain/APT-HLS-AI

\begin{table}
  \caption{Performance Comparison of Models}
  \label{table-1}
  \centering
  \begin{tabular}{ll}
    \toprule
    \cmidrule(r){1-2}
    Name  &  Score~(RMSE)  \\
    \midrule
    GPT4o & 13.79  \\
    GNN-DSE   & 10.00 \\
    LLAMA2  & 8.93   \\
    Fine tuned HARP  & 2.82  \\
    GPT4o+HARP & 6.60 \\
    Agentic-HLS & 4.21 \\    
    \bottomrule
  \end{tabular}
\end{table}

\section{Discussion}

Limitation of CoT prompting: The technique is less effective with smaller models. To achieve meaningful gains, it’s best to apply CoT in proportion to the model’s size, as smaller models (less than 100B parameter) may produce less coherent reasoning with CoT prompting.

We plan to use models such LLAMA3 to conduct this experiment however we were short of tie. Tat why we chose GPT4o. With emerge of larger models, we predict that agents will become more effective in reasoning and predicting design performances.
As future work, we plan to extend the reasoning with synthesiser version considerations to achieve more accurate results. 
We believe our Agentic-HLS approach once published can attract attention of agentic workflow researchers and designers to the Neurips 2024 AI for EDA community and we are keen to explore the opportunity with larger models such as GPT-4o1 capable of deep reasoning. 

\section{Acknowledgment}

We would like to thank to APT group member Zheyu Liu for his valuable efforts on this project. We would also like to thank Dr Pavlos Petoumenos of APT group of the  of Manchester for facilitating the use of A100 server for local deployments and Microsoft Azure for cloud credits to cover part of the costs of deploying GPT4o.

\section*{References}

\medskip

{
\small

[1] Yunsheng Bai, Atefeh Sohrabizadeh, Zongyue Qin, Ziniu Hu, Yizhou Sun, and Jason Cong. "Towards a Comprehensive Benchmark for FPGA Targeted High-Level Synthesis." NeurIPS, 2023.

[2] Qin, Zongyue, et al. "Cross-Modality Program Representation Learning for Electronic Design Automation with High-Level Synthesis." arXiv preprint arXiv:2406.09606 (2024).

[3] Sohrabizadeh, Atefeh, et al. "Robust GNN-based representation learning for HLS." 2023 IEEE/ACM International Conference on Computer Aided Design (ICCAD). IEEE, 2023.

[4] Sheikholeslam, S. A., \& Ivanov, A. (2024). SynthAI: A Multi Agent Generative AI Framework for Automated Modular HLS Design Generation. arXiv preprint arXiv:2405.16072.

[5] Pearce, H., Tan, B., \& Karri, R. (2020, November). Dave: Deriving automatically verilog from english. In Proceedings of the 2020 ACM/IEEE Workshop on Machine Learning for CAD (pp. 27-32).

[6] Chang, K., Wang, Y., Ren, H., Wang, M., Liang, S., Han, Y., ... \& Li, X. (2023). ChipGPT: How far are we from natural language hardware design. arXiv preprint arXiv:2305.14019.

[7] Wei, J., Wang, X., Schuurmans, D., Bosma, M., Xia, F., Chi, E., ... \& Zhou, D. (2022). Chain-of-thought prompting elicits reasoning in large language models. Advances in neural information processing systems, 35, 24824-24837.

[8] https://www.kaggle.com/competitions/machine-learning-contest-for-high-level-synthesis/

}
\end{document}